\documentclass[journal]{./sty/IEEEtran}

\pdfoutput=1

\ifCLASSINFOpdf
\else
\fi
\usepackage[cmex10]{amsmath}
\usepackage{amssymb}

\usepackage{graphicx}       
\usepackage{float}

\usepackage{algorithmic}

\usepackage{array}

\usepackage{fixltx2e}

\usepackage{stfloats}

\usepackage{url}
\usepackage{multirow}
\usepackage[justification=centering]{caption} 
\usepackage{booktabs}       

\makeatletter
\let\NAT@parse\undefined
\makeatother
\usepackage{hyperref}  

\begin{document}

\title{Point Cloud Semantic Segmentation using Multi-scale Sparse Convolution Neural Network}

\author{Yunzheng~Su,
        Jie~Cao,
        Lei~Jiang
\thanks{Yunzheng Su, Jie Cao, and Lei Jiang are with the School of Optics and Photonics, Beijing Institute of Technology, Beijing 100081, China(e-mail: suyunzzz@bit.edu.cn; caojie@bit.edu.cn; fabulousJerry@163.com).}
}

%
%

\markboth{Journal of \LaTeX\ Class Files,~Vol.~13, No.~9, September~2014}%
{Shell \MakeLowercase{\textit{et al.}}: Bare Demo of IEEEtran.cls for Journals}

\maketitle

\begin{abstract}
In recent years, with the development of computing resource and LiDAR, point cloud semantic segmentation has attracted many researchers. For the sparsity of point cloud, although there is already a way to deal with sparse convolution, multi-scale feature are not considered. In this letter, we propose a multi-scale feature fusion module based on multi-scale sparse convolution and an attentive channel feature filter module based on channel attention mechanism and build a point cloud segmentation network framework based on these modules. By introducing multi-scale feature fusion module and attentive channel feature filter module, the network could capture richer features, which improves the performance of point cloud semantic segmentation. Experimental results on Stanford large-scale 3-D Indoor Spaces(S3DIS) dataset and SemanticKITTI dataset demonstrate effectiveness and superiority of the proposed mothod.
\end{abstract}


%
\IEEEpeerreviewmaketitle

\section{Introduction}
\IEEEPARstart {W}{ith} the development of stereo matching algorithms and 3D sensors, point cloud is playing a significant role in several domains.
High-quality point cloud is the bridge between virtual world and real world. Through the processing of point cloud, we can perceive the environment clearly. For instance, semantic information can enrich the information of surrounding. Point cloud semantic segmentation is fundamental to some researchs such as computer vision, intelligent driving, remote sensing mapping, smart cities, etc.
Different from images, point cloud is unordered, unstructured, and have uneven density distribution at different distances. These characteristics make point cloud semantic segmentation more difficult than images. Nevertheless, research in point cloud semantic segmentation has seen an increase in the past few years, a series of point cloud semantic segmentation datasets have emerged, such as S3DIS \cite{armeni20163d}, semanticKITTI \cite{behley2019semantickitti}, etc.

Unstructured property and sparsity bring challenges to point cloud semantic segmentation, however, there are many successful methods introduced in review \cite{guo2020deep}. According to the different representations of point cloud, existing methods can be divided into four categories, including point-based, projection-based, voxel-based ,and hybrid methods. Although point-based methods can processing point cloud directly, it should be pointed out that point-based methods need to perform relatively complex preprocessing on the scene point cloud. In these mothods, scene point cloud needs to be divided into some regular cubes and sampled to generate training data(\cite{qi2017pointnet} \cite{qi2017pointnet++} \cite{zhao2019pointweb} \cite{zhang2019shellnet} \cite{lin2020adaptive}), which inevitably destroys the consistency of the point cloud. Projection-based methods (\cite{tatarchenko2018tangent}, \cite{lawin2017deep}, \cite{milioto2019rangenet++}, \cite{wu2019squeezesegv2}, \cite{tchapmi2017segcloud}, \cite{rethage2018fully}) are sensitive to occlusion and suffer from the loss of information.

Dense voxel-based methods(\cite{tchapmi2017segcloud, rethage2018fully}) voxelize the point cloud and directly apply standard 3D convolution for point cloud semantic segmentation. Although these methods keep the original dimension of the point cloud, the voxelization step inherently introduces discretization artifacts and loss of information. It is important to choose an appropriate grid resolution to have a trade-off between segmentation performance and efficiency. To this end, sparse convolution is proposed(\cite{yan2018second, graham20183d, choy20194d, tang2020searching}), which accelerates the computation speed by reducing useless computation at empty voxel in point cloud. However, affected by voxel-resolution, it is not easy to perceive small-scale objects.

To further leverage all available information, hybrid methods are proposed. PVCNN \cite{liu2019pvcnn} takes point-voxel representation as 3D input data to reduce memory consumption, while convolution in voxel reduces irregular sparse data access and improves locality, it use dense 2D convolution to extract feature, which leads to unnecessary calculation. 
It is difficult for dense voxel-based mothods to use high resolution voxel to get more fine-grained features limited by computational complexity. Meanwhile, sparse voxel-based mothods can only extract single-scale feature or can not fusion multi-scale feature effectively. Thus, compared with single-scale sparse voxel-based methods, multi-scale feature is crucial to perceive small objects \cite{tang2020searching,cheng20212}.

In this letter, we propose a multi-scale feature fusion module based on sparse convolution and an attentive channel feature filter module based on channel attention mechanism. By introducing multi-scale sparse convolution, the network could capture richer features based on convolution kernel with different sizes, improving the segmentation performance of point cloud. Experimental results on Stanford 3D Indoor Semantics (S3DIS) Dataset \cite{armeni20163d} and SemanticKITTI \cite{behley2019semantickitti}, demonstrate effectiveness and superiority of the proposed mothod. Overall, our key contributions are:
\begin{itemize}
  \item We propose a novel multi-scale feature fusion module(MFFM) based on sparse convolution, and an attentive channel feature filter module(ACFFM).
  \item We construct a hierarchical network based on these modules, which achieves comparative semantic segmentation performance on Stanford 3D Indoor Semantics (S3DIS) Dataset and SemanticKITTI.
\end{itemize}

\section{Proposed method}
\subsection{Multi-scale Feature Fusion Module}
In the two-dimensional convolutional neural network, in order to capture rich neighborhood information, multi-scale feature fusion method is used. Inspired by Fang et al. \cite{fang2019pyramid}, this letter designs a multi-scale feature fusion module, which uses convolution kernel of different sizes to extract feature of different scales in the point cloud to achieve a complementary effect. 

As shown in Fig. \ref{pic:architecture}, this module utilizes multi-scale sparse convolution of large-scale, medium-scale, small-scale, and point-level scale in parallel, and captures features of different scales as $x_1$, $x_2$, $x_3$, $x_0$ respectively. Among them, the convolution of the point scale level has a kernel size of 1, which means that only single point feature transformation is performed. For multi-scale features, basic fusion method is direct addition, however, features are regarded as same weight, the features of different scales should have different influences. As shown in Fig. \ref{pic:architecture}, inspired by SKNet \cite{li2019selective}, we propose a multi-scale feature fusion module. Firstly, the large-scale, medium-scale ,and small-scale features are added to obtain intermediate feature map $X$, and then three different $MLP$s are used to perform three transformations on the intermediate feature map $X$:
\begin{equation} \label{Multi Scale Feature Fusion 1}
  \begin{split}
      f:{X}\rightarrow{U}\in\mathbb{R}^{N\times4} \\
\end{split}
\end{equation}

In the formula \ref{Multi Scale Feature Fusion 1}, each transformation $f_i$ is a combination of convolution layer, activation function layer and $BN$ (BatchNormalization) layer. The obtained features of different scales are concatenated together and sent to the $Softmax$ activation function to obtain the corresponding score of each scale:
\begin{gather} 
    {U_{cat}}=cat({U_1}, {U_2}, {U_3})\label{Fusion2} \\
    {S}=softmax({U_{cat}})\label{Fusion3}
\end{gather}

In the formula \ref{Fusion2}, ${U_{cat}}\in{\mathbb{R}^{N\times3}}$ represents the feature map after concatenated by formula \ref{Fusion2}, and in the formula \ref{Fusion3}, ${S}\in{\mathbb{R}^{N\times4}}$ represents the attention score of each scale.
Finally, multi-scale features are added weightedly according to corresponding score to obtain the final multi-scale fusion feature. In order to retain the point-wise feature, the point-level feature after a Submanifold Sparse Convolution with a kernel size of 1 is added to weighted multi-scale feature and make a transformation:
\begin{equation}
    {O} = {F}\left(x_{0}+x_{1}\cdot{S}[:, 0]+x_{2} \cdot {S}[:, 1]+x_{3}\cdot {S}[:, 2]\right)\label{SKNet:output}
\end{equation}
where $x_0$ is the output of the Submanifold Sparse Convolution with kernel size 1, $F(\cdot)$ is the combination of sparse convolution and BatchNormalization.

\subsection{Attentive Channel Feature Filter Module}
Fig. \ref{pic:architecture} shows the attentive channel feature filter module based on channel attention mechanism. For the point cloud feature map of different scales passed by the multi-scale feature fusion module, processing ${x_1}$, ${x_2}$, ${x_3}$ through the residual unit firstly, the outputs are recorded as ${x_1^{\prime}}$, ${x_2^{\prime}}$, ${x_3^{\prime}}$, then connect ${x_1^{\prime}}$, $x_2^{\prime}$, $x_3^{\prime}$ and introduce nonlinearity. In the SE Module, the input point cloud feature map is sent to global pooling operation to obtain global feature:
\begin{equation} \label{Sequeeze}
    x_{c} =\frac{1} {N}{\sum_{i=1}^{N} x_i}
\end{equation}
where $N$ is the voxel number, $x_i\in{\mathbb{R}^{N\times{C}}}$ donates voxel feature.

For the global feature of the point cloud after the Sequeeze operation \ref{Sequeeze}, it is necessary to further obtain the feature relationship between different channels. This step is composed of two linear transformations followed by activation function:
\begin{equation} \label{extract}
    s = {\sigma}(W_{2}\delta(W_{1}x_{c}))
\end{equation}
where $\delta$ is $ReLU$ activate function, $\sigma$ is $Sigmoid$ activate function. $W_1\in{\mathbb{R}^{{\frac{C}{r}}\times{C}}}$ is the first linear transformation, compressing the number of channel from $C$ to $\frac{C}{r}$ and reducing the complexity of the model, where $r$ is the dimension reduction scale factor. $W_2\in{\mathbb{R}^{{C}\times{\frac{C}{r}}}}$ is the second linear transformation which restores original number of channels.

Finally, the channel attention coefficients obtained by formula \ref{extract} are weighted on all channels:
\begin{equation} \label{scale}
    x_{output}=s_c\cdot{x_c}
\end{equation}
where $s_c$ and $x_c$ represent the weight score and feature of different channels, $s_{output}$ represent the output of ACFFM. 

\subsection{Overall architecture}
\begin{figure*}[]
  \centering
  \includegraphics[width=\textwidth,height=0.4\textwidth]{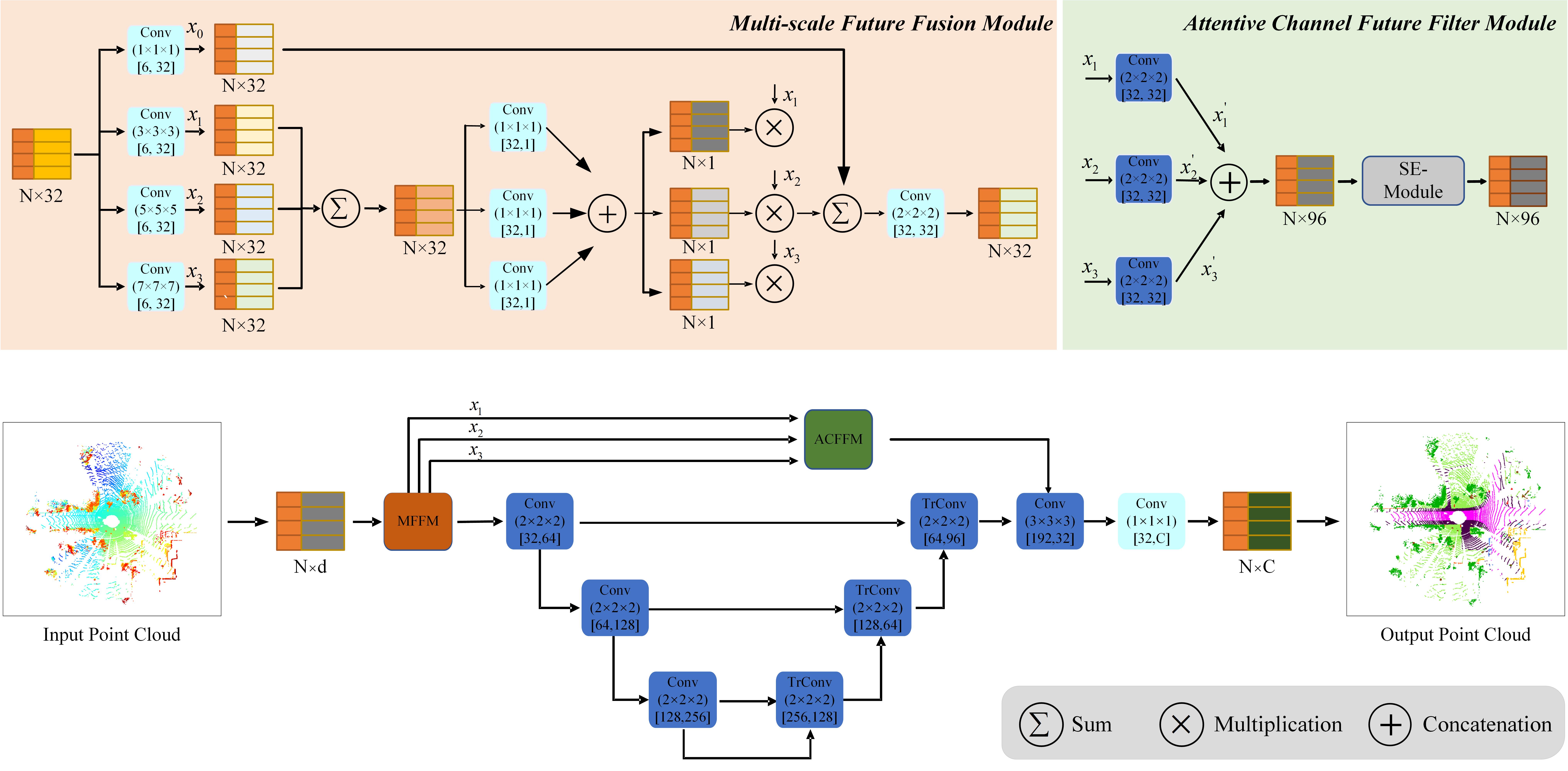}
  \caption{Proposed MSSNet architecture}
  \label{pic:architecture}
\end{figure*}

The basic network of this letter refers to the hierarchical structure of PointNet++, and the network consists of encoders and decoders(Fig. \ref{pic:architecture}). The difference is that PointNet++ is a point-based point cloud segmentation model. The $FPS$(farthest point sampling) algorithm is used in downsampling module which has a time complexity of $O(n^2)$. MSSNet is a voxel-based point cloud segmentation method. The voxelization based on hash table is used by the downsampling module which has time complexity of $O(1)$. In the neighborhood building module, PointNet++ uses $KNN$(K-Nearest Neighbor searching) or spherical neighborhood retrieval, the time complexity is $O(n)$. This letter adopt a neighborhood building method based on hash table, and the time complexity is $O(1)$. Consequently, the basic network in this letter is faster than PointNet++.

\subsection{Loss Function}
We leveraged a linear combination of Cross-Entroy loss and lov{\'a}sz-softmax loss \cite{berman2018lovasz} to optimize our network:

\begin{equation}\label{equ:loss}
  L(y, \hat{y}) = w_{ce}L_{ce}(y, \hat{y}) + w_{lovasz}L_{lovasz}(y, \hat{y})
\end{equation}
where $w_{ce}$ and $w_{lovasz}$ denote the weight of Cross-Entroy loss and Lov{\'a}sz loss respectively. They are both set as 1 in our experiments.

\section{EXPERIMENTS}

\subsection{Implementation details}
We implement the proposed MSSNet with Pytorch. SGD optomizer with momentum is used with an initinal learning rate of 0.24 in our experiments. The voxel size is 0.05m. $w_{ce}$ and $w_{lovasz}$ in combined loss function are both set to 1.0. Data augmentation methods are consistent with the paper \cite{choy20194d}, including random scaling, rotation around the Z axis, spatial translation, and random jittering. All our experiments are conducted on two NVIDIA RTX2080Ti GPUs.

\subsection{Semantic Segmentation on Public Benchmarks}
\subsubsection{Evaluation on S3DIS}
S3DIS \cite{armeni20163d} dataset is a large-scale indoor point cloud scene dataset created by Stanford University in 2016. S3DIS dataset consists of 271 rooms belonging to 6 large areas. Each point in this dataset contains 3-D coordinate, RGB information, and is labeled as one of thirteen indoor semantic categories.

Table \ref{table:s3dis eval} shows quantitative result of previous methods and MSSNet. From Table \ref{table:s3dis eval}, we can observe that MSSNet outperforms other methods in both mIoU and OA, owing to MSSNet can extract and fusion multi-scale feature adaptively. It should be noted that some methods in Table \ref{table:s3dis eval} divide point cloud into blocks(\cite{qi2017pointnet} \cite{qi2017pointnet++} \cite{li2018pointcnn}), which destroys consistency of the point cloud inevitably. Conversely, MSSNet regards the entire room as input, for some objects, long-distance context information could be better perceived.
\begin{table*}[]
  \caption{Segmentation results of different networks on the S3DIS dataset}
  \centering
  \resizebox{\textwidth}{!}{%
  \begin{tabular}{cccccccccccccccc}
  \hline
  Method          & Ceiling & Floor & Wall & Beam & Clmn & Window        & Door & Chair & Table         & Bkcase & Sofa          & Board & Clutter & mIoU & mAcc \\ \hline
  PointNet \cite{qi2017pointnet}        & 87.4    & 97.8  & 71.2 & 0.0  & 9.2  & 52.1          & 16.3 & 48.6  & 58.2          & 48.3   & 3.2           & 39.0  & 36.2    & 43.7 & 52.6 \\
  SegCloud \cite{tchapmi2017segcloud}        & 90.1    & 96.1  & 69.9 & 0.0  & 18.4 & 38.4          & 23.1 & 75.9  & 70.4          & 58.4   & 40.9          & 13.0  & 41.6    & 48.9 & 57.4 \\
  TangentConv \cite{tatarchenko2018tangent}     & 90.5    & 97.7  & 74.0 & 0.0  & 26.7 & 39.0          & 31.3 & 77.5  & 69.4          & 57.3   & 38.5          & 48.8  & 39.8    & 52.8 & 60.7 \\
  3D RNN \cite{ye20183d} &
    \textbf{95.2} &
    \textbf{98.6} &
    77.4 &
    \textbf{0.8} &
    9.8 &
    52.7 &
    27.9 &
    76.8 &
    78.3 &
    58.6 &
    27.4 &
    39.1 &
    51.0 &
    53.4 &
    71.3 \\
  PointNet++ \cite{qi2017pointnet++}      & 90.7    & 97.0  & 75.9 & 0.0  & 6.3  & \textbf{58.3} & 19.4 & 74.9  & 69.5          & 62.2   & 51.0          & 57.4  & 42.9    & 54.4 & 64.2 \\
  PointCNN \cite{li2018pointcnn}        & 92.3    & 98.2  & 79.4 & 0.0  & 17.6 & 22.8          & 62.1 & 80.6  & 74.4          & 66.7   & 31.7          & 62.1  & 56.7    & 57.3 & 63.9 \\
  SuperpointGraph \cite{landrieu2018large} & 89.4    & 96.9  & 78.1 & 0.0  & 42.8 & 48.9          & 61.6 & 84.7  & \textbf{75.4} & 69.8   & 52.6          & 2.1   & 52.2    & 58.0 & 66.5 \\
  PCCN \cite{wang2018deep}            & 90.3    & 96.2  & 75.9 & 0.3  & 6.0  & 69.5          & 63.5 & 66.9  & 65.6          & 47.3   & \textbf{68.9} & 59.1  & 46.2    & 58.3 & 67.0 \\
  MinkowskiNet \cite{choy20194d} &
    92.6 &
    98.2 &
    \textbf{84.3} &
    0 &
    \textbf{37.6} &
    51.7 &
    \textbf{73.8} &
    \textbf{90.2} &
    73.8 &
    69.7 &
    63.7 &
    66.8 &
    55.3 &
    65.4 &
    72.3 \\
  MSSNet(Ours) &
    91.0 &
    96.2 &
    83.6 &
    0 &
    28.9 &
    56.1 &
    72.7 &
    82.4 &
    74.5 &
    \textbf{73.1} &
    67.8 &
    \textbf{76.5} &
    \textbf{59.9} &
    \textbf{67.0} &
    \textbf{74.9} \\ \hline
  \end{tabular}%
  }
  \label{table:s3dis eval}
  \end{table*}

\begin{figure}
  \centering
  \includegraphics[width=0.5\textwidth,height=0.4\textwidth]{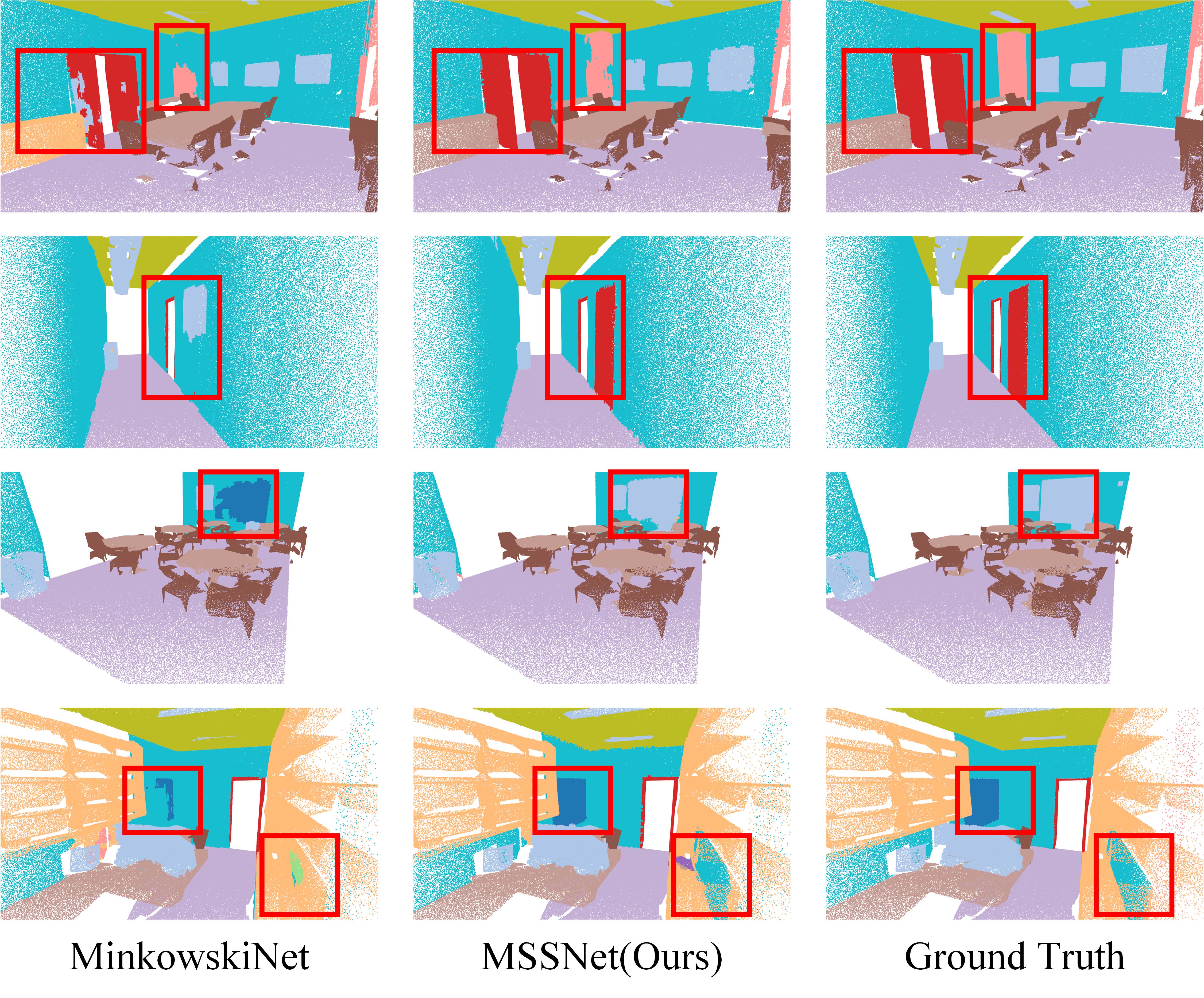}
  \caption{Examples semantic segmentation results on S3DIS. We viuslize and compare our segmentation results with MinkowskiNet. Different colors denote different categories of the object in the scene.}
  \label{pic:s3dis}
\end{figure}

Fig. \ref{pic:s3dis} visualizes some semantic segmentation results on Area 5 of S3DIS \cite{armeni20163d}. As we can see,  due to the well-designed multi-scale feature scheme, our MSSNet performs better segmentation result of fine-grained details. For instance, the door in first row and second row, the clutter in the third row, and the board in the last row.

\subsubsection{Evaluation on SemanticKITTI}
SemanticKITTI \cite{behley2019semantickitti} is a dataset with point-wise semantic annotations built by the University of Bonn on the basis of the KITTI dataset benchmark. The dataset includes 22 sequences, with a total of 43551 frames of lidar data. Among the driving sequences, 10 sequences(00 to 07, 09, 10) are used to training, and sequence 08 was for evaluation. The remaining sequences are used for the comparison of our model's performance with previous methods. Comparing with the S3DIS dataset, the density of the SemanticKITTI dataset is more uneven.

\begin{table*}[htb]
  \caption{SemanticKITTI test leaderboard results}
  \centering
  \resizebox{\textwidth}{!}{%
  \begin{tabular}{ccccccccccccccccccccc}
  \hline
  Method &
    car &
    bicycle &
    motorcycle &
    truck &
    other-vehicle &
    person &
    bicyclist &
    motorocyclist &
    road &
    parking &
    sidewalk &
    other-ground &
    building &
    fence &
    vegetation &
    trunk &
    terrain &
    pole &
    traffic-sign &
    mIoU \\ \hline
  PointNet \cite{qi2017pointnet} &
    46.3 &
    1.3 &
    0.3 &
    0.1 &
    0.8 &
    0.2 &
    0.2 &
    0.0 &
    61.6 &
    15.8 &
    35.7 &
    1.4 &
    41.4 &
    12.9 &
    31.0 &
    4.6 &
    17.6 &
    2.4 &
    3.7 &
    14.6 \\
  RandLANet \cite{hu2020randla} &
    94.2 &
    26.0 &
    25.8 &
    40.1 &
    38.9 &
    49.2 &
    48.2 &
    7.2 &
    90.7 &
    60.3 &
    73.7 &
    20.4 &
    86.9 &
    56.3 &
    81.4 &
    61.3 &
    66.8 &
    49.2 &
    47.7 &
    53.9 \\
  KPConv \cite{thomas2019kpconv} &
    96.0 &
    30.2 &
    42.5 &
    33.4 &
    44.3 &
    61.5 &
    61.6 &
    11.8 &
    88.8 &
    61.3 &
    72.7 &
    31.6 &
    90.5 &
    64.2 &
    84.8 &
    69.2 &
    69.1 &
    56.4 &
    47.4 &
    58.8 \\
  SequeezeSegV3 \cite{xu2020squeezesegv3} &
    92.5 &
    38.7 &
    36.5 &
    29.6 &
    33.0 &
    45.6 &
    46.2 &
    20.1 &
    91.7 &
    63.4 &
    74.8 &
    26.4 &
    89.0 &
    59.4 &
    82.0 &
    58.7 &
    65.4 &
    49.6 &
    58.9 &
    55.9 \\
  RangeNet++ \cite{milioto2019rangenet++} &
    91.4 &
    25.7 &
    34.4 &
    25.7 &
    23.0 &
    38.3 &
    38.8 &
    4.8 &
    91.8 &
    65.0 &
    75.2 &
    27.8 &
    87.4 &
    58.6 &
    80.5 &
    55.1 &
    64.6 &
    47.9 &
    55.9 &
    52.2 \\
  SalsaNet \cite{aksoy2020salsanet} &
    91.9 &
    48.3 &
    38.6 &
    38.9 &
    31.9 &
    60.2 &
    59.0 &
    19.4 &
    91.7 &
    63.7 &
    75.8 &
    29.1 &
    90.2 &
    64.2 &
    81.8 &
    63.6 &
    66.5 &
    54.3 &
    62.1 &
    59.5 \\
  PolarNet \cite{zhang2020polarnet} &
    93.8 &
    40.3 &
    30.1 &
    22.9 &
    28.5 &
    43.2 &
    40.2 &
    5.6 &
    90.8 &
    61.7 &
    74.4 &
    21.7 &
    90.0 &
    61.3 &
    84.0 &
    65.5 &
    67.8 &
    51.8 &
    57.5 &
    54.3 \\
  MinkowskiNet \cite{choy20194d} &
    - &
    - &
    - &
    - &
    - &
    - &
    - &
    - &
    - &
    - &
    - &
    - &
    - &
    - &
    - &
    - &
    - &
    - &
    - &
    63.1 \\
  FusionNet \cite{zhang2020deep} &
    95.3 &
    47.5 &
    37.7 &
    41.8 &
    34.5 &
    59.5 &
    56.8 &
    11.9 &
    \textbf{91.8} &
    68.8 &
    \textbf{77.1} &
    30.8 &
    \textbf{92.5} &
    \textbf{69.4} &
    84.5 &
    69.8 &
    68.5 &
    60.4 &
    66.5 &
    61.3 \\
  TornadoNet \cite{gerdzhev2021tornado} &
    94.2 &
    55.7 &
    48.1 &
    40.0 &
    38.2 &
    63.6 &
    60.1 &
    34.9 &
    89.7 &
    66.3 &
    74.5 &
    28.7 &
    91.3 &
    65.6 &
    85.6 &
    67.0 &
    \textbf{71.5} &
    58.0 &
    65.9 &
    63.1 \\
  AMVNet \cite{liong2020amvnet} &
    96.2 &
    \textbf{59.9} &
    \textbf{54.2} &
    48.8 &
    45.7 &
    \textbf{71.0} &
    65.7 &
    11.0 &
    90.1 &
    \textbf{71.0} &
    75.8 &
    \textbf{32.4} &
    92.4 &
    69.1 &
    \textbf{85.6} &
    71.7 &
    69.6 &
    62.7 &
    \textbf{67.2} &
    65.3 \\
  SPVCNN \cite{tang2020searching} &
    - &
    - &
    - &
    - &
    - &
    - &
    - &
    - &
    - &
    - &
    - &
    - &
    - &
    - &
    - &
    - &
    - &
    - &
    - &
    63.8 \\
  MSSNet(Ours) &
    \textbf{96.8} &
    52.2 &
    48.5 &
    \textbf{54.4} &
    \textbf{56.3} &
    67.0 &
    \textbf{70.9} &
    \textbf{49.3} &
    90.1 &
    65.5 &
    74.9 &
    30.2 &
    90.5 &
    64.9 &
    84.9 &
    \textbf{72.7} &
    69.2 &
    \textbf{63.2} &
    65.1 &
    \textbf{66.7} \\ \hline
  \end{tabular}%
  }
  \label{tab:SemanticKITTI test}
  \end{table*}

Table \ref{tab:SemanticKITTI test} is quantitative result of experiment on SemanticKITTI test data, compared models contain point-based method, projection-based mothod, voxel-based mothod and hybrid mothod, MSSNet belong to voxel-based method, the results of MinkowskiNet and SPVCNN come from the paper \cite{tang2020searching}. It should be pointed out that other methods in Table \ref{tab:SemanticKITTI test} may use some additional tricks to improve IoU, while the method in this letter does not use any additional tricks. MSSNet has achieved leading results in mIoU and IoU of some small-scale objects(such as bicycles, motorcyclists, poles, etc.).

\begin{figure}[htp]
  \centering
  \includegraphics[width=0.5\textwidth,height=0.3\textwidth]{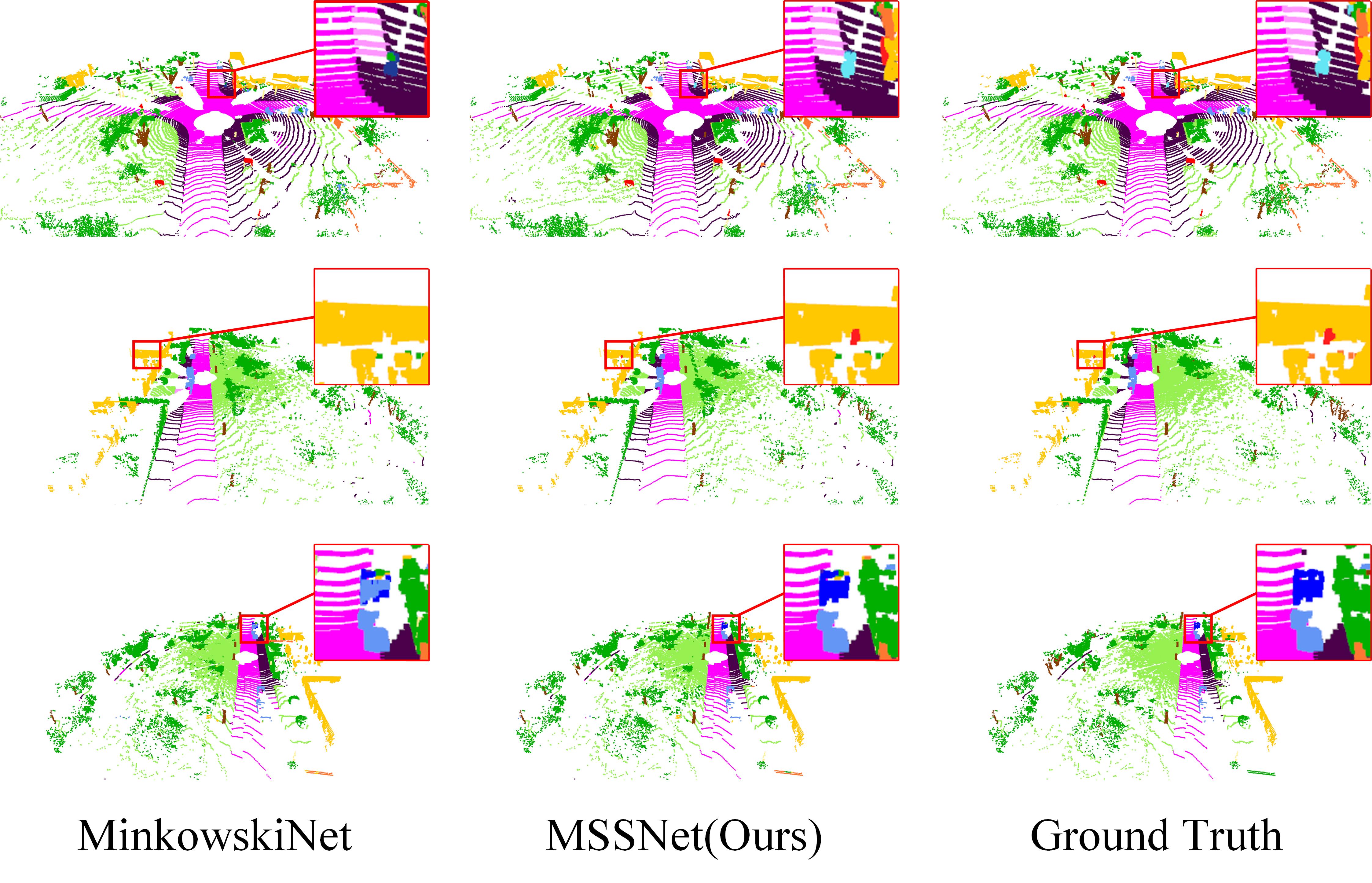}
  \caption{Examples semantic segmentation results on SemanticKITTI. We viuslize and compare our segmentation results with MinkowskiNet. Different colors denote different categories of the object in the scene.}
  \label{pic:SemanticKITTI}
\end{figure}

Fig. \ref{pic:SemanticKITTI} visualizes the semantic segmentation results of MSSNet. The test datasets is sequence 08 in SemanticKITTI \cite{behley2019semantickitti}. As we can see, MinkowskiNet is prone to wrong segmentation for some small-scale objects, MSSNet benefits from multi-scale feature, the ability to perceive small-scale objects is improved. For instance, The bicycle in the red square in first row is segmented into other categories by MinkowskiNet, but MSSNet can achieve correct semantic segmentation.

\subsection{Ablation Study}
In order to verify contribution of each component in our network, several groups ablation experiments are designed on S3DIS dataset and SemanticKITTI dataset. The detailed results can be found in Table \ref{tab:Ablation}. 
\begin{table}[H]
  \caption{Ablation study of the proposed method}
  \centering
  \resizebox{0.5\textwidth}{!}{%
  \normalsize
  \begin{tabular}{cccccc}
  \hline
  Module                    & MFFM & ACFFM & lov{\'a}sz & S3DIS(\%)  & SemanticKITTI(\%)     \\ \hline
  Baseline                  &      &     &                  & 64.33      & 64.03    \\
  \multirow{4}{*}{Proposed} & \checkmark   &     &                  & 66.35      & 65.48    \\
                            &      & \checkmark  &                  & 64.50    & 65.13       \\
                            & \checkmark   & \checkmark  &                  & 66.47     & 66.23     \\
                            & \checkmark   & \checkmark  & \checkmark               & \textbf{67.03} & \textbf{66.95} \\ \hline
  \end{tabular}%
  }
  \label{tab:Ablation}
  \end{table}

From Table \ref{tab:Ablation} we can see that the multi-scale feature fusion module and the attentive channel feature filter module proposed in this letter can enrich the multi-scale feature representation to improve the semantic segmentation performance. To further validate the effectiveness of Lovasz-Softmax loss, We implement our method with cross-entropy loss, the combination of cross-entropy loss and Lovasz-Softmax loss, respectively. As shown in Table \ref{tab:Ablation} the combination of two loss improves the overall segmentation performance, because this loss can directly optimize the mIoU performance.

\subsection{Evaluation by distance}
\begin{figure}[H]
  \centering
  \includegraphics[width=0.5\textwidth,height=0.3\textwidth]{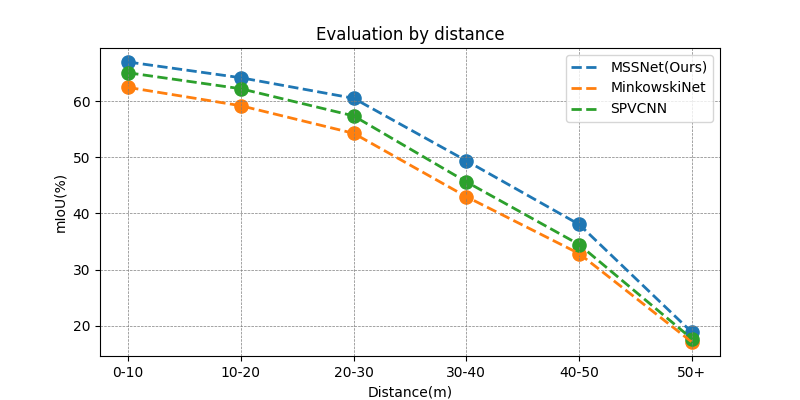}
  \caption{mIoU vs Distance}
  \label{pic:analysis}
\end{figure}

In order to verify the robustness of MSSNet, experiment is conducted on SemanticKITTI validation set. Fig. \ref{pic:analysis} illustrates the mIoU of MinkowskiNet, SPVCNN and MSSNet(proposed) with different distances. In different regions, the mIoU value of MSSNet is always higher than SPVCNN and MinkowskiNet, which MinkowskiNet is the lowest, this is because the SPVCNN utilizes fine-grained point-wise feature, so higher mIoU can be obtained with low density at longer distance. MSSNet can capture multi-scale feature, therefore, it has better segmentation result in each distance range, it is proved that the mothod we proposed has good robustness with point cloud density.


\section{CONCLUSION}
In this letter, we propose multi-scale feature fusion module and attentive channel feature filter module, which enrich multi-scale feature in point cloud semantic segmentation. Based on these modules, we construct a hierarchical network called MSSNet for point cloud semantic segmentation task. MSSNet achieves comparative performance compared with classical methods on S3DIS dataset and SemanticKITTI dataset, which demonstrate the effectiveness and superiority.

\ifCLASSOPTIONcaptionsoff
  \newpage
\fi

\bibliographystyle{IEEEtran}
\bibliography{reference}

\end{document}